\documentclass{article}
\usepackage{spconf,amsmath,graphicx}
\usepackage{yhmath}
\usepackage{booktabs}
\usepackage{multirow}

\title{ATCA: AN ARC TRAJECTORY BASED MODEL WITH CURVATURE ATTENTION FOR Video FRAME INTERPOLATION}
\name{Jinfeng Liu \qquad Lingtong Kong \qquad Jie Yang\thanks{Thanks to the supporting of NSFC, China (No: 61876107, U1803261).}}
\address{Institute of Image Processing and Pattern Recognition, Shanghai Jiao Tong University, China}
\begin{document}
%
\maketitle
\begin{abstract}
Video frame interpolation is a classic and challenging low-level computer vision task. Recently, deep learning based methods have achieved impressive results, and it has been proven that optical flow based methods can synthesize frames with higher quality. However, most flow-based methods assume a line trajectory with a constant velocity between two input frames. Only a little work enforces predictions with curvilinear trajectory, but this requires more than two frames as input to estimate the acceleration, which takes more time and memory to execute. To address this problem, we propose an arc trajectory based model (ATCA), which learns motion prior from only two consecutive frames and also is lightweight. Experiments show that our approach performs better than many SOTA methods with fewer parameters and faster inference speed.
\end{abstract}
\section{Introduction}
\label{sec:intro}

Video frame interpolation (VFI) is a classic low-level task of video enhancement in computer vision, which aims to synthesize one or more intermediate frames between the consecutive input frames. It has many practical applications, such as frame rate conversion, video editing, motion deblurring, inter-frame compression and medical imaging.

Recently, deep learning based VFI methods can be categorized as kernel-based methods and flow-based methods. Long \emph{et al}. \cite{Learning2016} firstly introduce a CNN model which directly estimates the intermediate frame using two input frames.
\begin{figure}[htb]
    \centering
    \includegraphics[width=8cm]{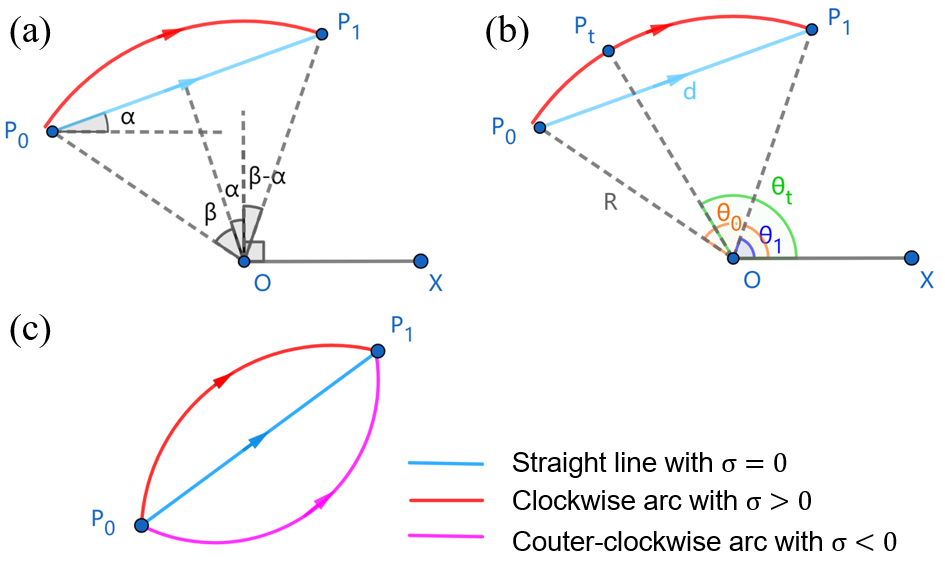}
    \caption{The geometric schematic of the arc trajectory model.}
    \label{f1}
\end{figure}
However, it usually causes blurry results. To address this problem, Niklaus \emph{et al}. \cite{niklaus2017} apply kernel-based methods AdaConv and SepConv, which generate the intermediate frame by locally convolving the input frames. These methods focus on where to find the output pixel from the input frames. Besides, AdaCoF \cite{adacof2020} proposes a novel operation, called adaptive collaboration of flows, that can refer to any number of pixels and any location. The second flow-based methods are to estimate the motion flows between the the input frames using optical flow algorithms, warp the input frames and synthesize the target frame guided by the warped results. Liu \emph{et al.} \cite{dvf2017} propose Deep Voxel Flow (DVF), to estimate a flow map directly pointing to reference locations. Jiang \emph{et al.} \cite{slomo2018} propose Super-Slomo, which combines an U-Net flow estimator and kernel-based image synthesis with the warped results. After that, more advanced flow estimation models like PWC-Net \cite{pwc2018}, which uses coarse-to-fine architecture to iteratively refine the flow map, are applied in most frame interpolation pipelines \cite{soft2020,con2018,qvi2019,Kong_2022_CVPR}. 
\begin{figure*}[htb]
    \centering
    \includegraphics[width=17cm]{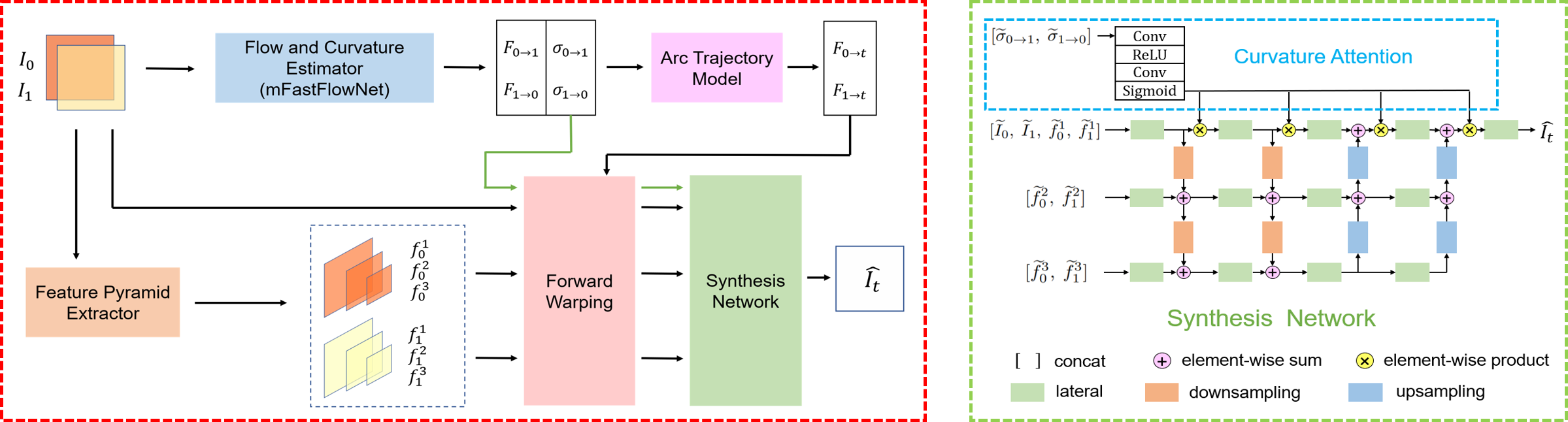}
    \caption{The left part is the overview pipeline of our ATCA model. The right part is the detailed structure of proposed synthesis network, where the lateral, downsampling and upsampling blocks refer to \cite{con2018}.}
    \label{f2}
\end{figure*}

However, most of these approaches mentioned above are based on such an assumption that a pixel between consecutive frames moves along a straight line at a constant speed. Although Xu \emph{et al.} \cite{qvi2019} propose QVI, an acceleration-aware algorithm, that allows predictions with curvilinear trajectory and variable velocity, it takes four consecutive frames as input to estimate acceleration information, which is time consuming, and can not learn motion trajectory adaptively from the frame context. To alleviate this problem, we propose an \textbf{A}rc \textbf{T}rajectory based frame interpolation model with \textbf{C}urvature \textbf{A}ttention (ATCA) in this paper. Overall, our major contributions are summarized as follows:~1) we propose ATCA to interpolate video frames, which takes two consecutive frames and assumes an arc trajectory between them, with the flow maps and curvature maps obtained by the joint estimator;~2) we introduce curvature attention into the synthesis network to improve the synthesis quality using the curvature maps;~3) our model is lightweight and fast, and outperforms many SOTA VFI methods.

\section{Proposed Method} 

With two consecutive frames $I_0$ and $I_1$, frame interpolation gives a prediction of the intermediate frame, denoted as $\hat{I_t}$, to reconstruct the ground truth $I_t$ as accurate as possible, where $t \in (0,1)$ denotes the arbitrary temporal position. To get high quality intermediate frame, we establish our model based on SoftSplat \cite{soft2020}. SoftSplat first estimates the bidirectional flows $F_{0 \rightarrow1}$ and $F_{1 \rightarrow0}$ between the input frames using an off-the-shelf optical flow method like PWC-Net \cite{pwc2018}. And then it employs softmax splatting, a form of forward warping, to warp $I_0$ according to $F_{0 \rightarrow t}=t \cdot F_{0 \rightarrow 1}$ and $I_1$ according to $F_{1 \rightarrow t}=(1-t) \cdot F_{1 \rightarrow 0}$. Pyramid features extracted from $I_0$ and $I_1$ are also warped to the temporal position $t$. Then, these intermediate results are fed into the synthesis network to obtain the final $\hat{I_t}$. To make the intermediate flows $F_{0 \rightarrow t}$ and $F_{1 \rightarrow t}$ more accurate, we model the motion trajectory of a pixel from $I_0$ to $I_1$ as an inferior arc and assume the velocity is constant, which is different from the linear model in SoftSplat. Also unlike QVI \cite{qvi2019}, we only utilize two consecutive frames to synthesize the intermediate frame and propose a novel joint estimator with optical flow, to learn the prior information about the curvature of the arc trajectory at each pixel. The overview of our model is shown in Fig. \ref{f2}.

\subsection{Arc Trajectory Model}
The arc trajectory model is depicted in Fig. \ref{f1}, where point $P_0$ and point $P_1$ denote the position of a pixel in the two input frames $I_0$ and $I_1$ respectively, with $\overline{{P_0 P_1}}^x = F_{0 \rightarrow 1}^x$ and $\overline{{P_0 P_1}}^y = F_{0 \rightarrow 1}^y$. And $\wideparen{P_0 P_1}$ is the arc trajectory with the center point $O$, the radius $R$ and the arc angle $\angle{P_0 O P_1 }=2 \beta$, while $\alpha$ is the dip angle of $\overline{{P_0 P_1}}$. So we have the following geometric relations
\begin{equation}
\label{eq1}
\begin{split}
    &d=\overline{{P_0 P_1}}=\sqrt{{F_{0 \rightarrow 1}^x}^2+{F_{0 \rightarrow 1}^y}^2}, \\
    &\alpha=atan2(F_{0 \rightarrow 1}^y,F_{0 \rightarrow 1}^x),~~~~~~~\beta=\arcsin{\frac{d}{2R}},
\end{split}
\end{equation}

\begin{equation}
\label{eq2}
\begin{split}
    &\theta_0=\alpha+\frac{\pi}{2}+\beta,~~~~~~~~~~~~~~\theta_1=\alpha+\frac{\pi}{2}-\beta, \\
    &\theta_t=(\theta_1-\theta_0)t+\theta_0=-2\beta t + \alpha+\frac{\pi}{2}+\beta,
\end{split}
\end{equation}
where $\theta_0$ and $\theta_1$ denote the polar angle of $P_0$ and $P_1$ in polar coordinate system respectively, while $\theta_t$ denotes the polar angle of $P_t$ in the intermediate frame $I_t$ and is calculated under the constant velocity assumption.

\subsection{Flow and Curvature Joint Estimator}
Although there exist some advanced optical flow models such as PWC-Net \cite{pwc2018}, LiteFlowNet \cite{lite2018} and so on, we choose a more efficient model, FastFlowNet \cite{ffn2021}, for fast and accurate optical flow prediction, which consists of three parts: the head enhanced pooling pyramid feature extractor, the center dense dilated correlation layer and the efficient shuffle block decoder. It takes in two consecutive frames and generates six pyramid levels to get multi-scale coarse-to-fine flows, while we remove the coarsest level ($\frac{1}{64}$ resolution) in this paper since the top five levels are enough. Besides, the channel number of the flow decoder at every scale is increased to 3. The first two channels are corresponding to the horizontal and vertical flows as usual, while the additional channel is used to estimate the curvature information. Let $\sigma=\sin \beta=\frac{d}{2R}$, we use $\sigma$ to measure the curvature, which is scale-invariant. So the additional channel is fed into a \texttt{tanh} layer, the output value of which is restricted in $[-1,1]$, to estimate the value of $\sigma$ for every pixel, forming a $\sigma$-map. The modified FastFlowNet is named mFastFlowNet in this paper and we use $h(\cdot)$ to represent it. With the two input frames $I_0$ and $I_1$, we have
\begin{equation}
\label{eq3}
\begin{split}
    &F_{0 \rightarrow 1}^x, ~F_{0 \rightarrow 1}^y, ~\sigma_{0 \rightarrow 1} = h(I_0, I_1),\\
    &R=\frac{d}{2\sigma_{0 \rightarrow 1}},~~~~~~~~~\beta=\arcsin{\sigma_{0 \rightarrow 1}}.
\end{split}
\end{equation}
\begin{figure*}[htb]
    \centering
    \includegraphics[width=17cm]{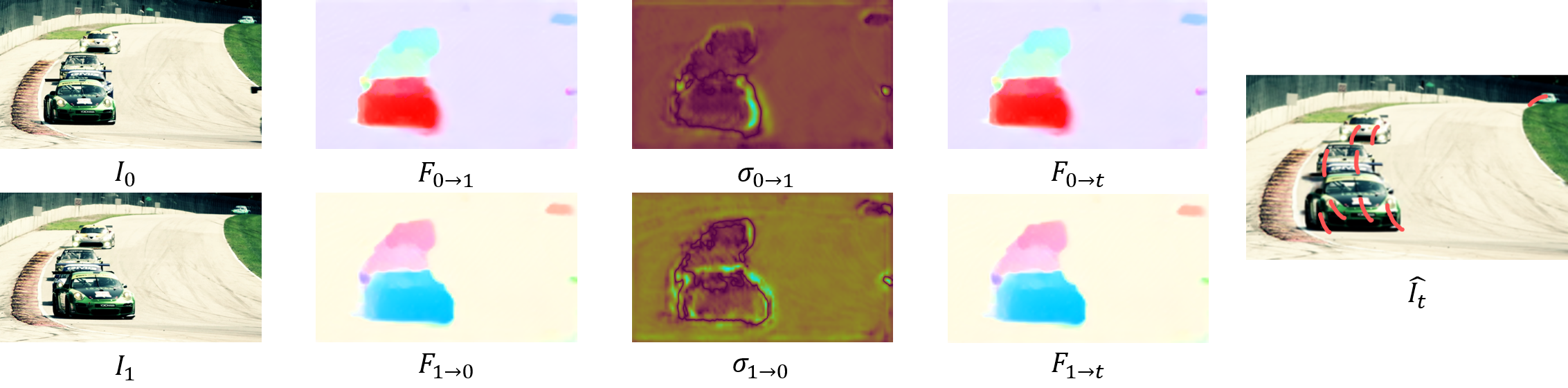}
    \caption{Visualization of the forward computation process of ATCA. The arc trajectory obtained by ATCA is drawn in $\hat{I_t}$.}
    \label{f3}
\end{figure*}
Note that too small $\sigma(0)$ means too large $R(\infty)$, which may cause numerical instability and gradient explosion in the training stage. Actually, the arc degenerates into a line when $\sigma=0$, hence we set a threshold to enforce the linear trajectory when $\sigma$ is too small. Then, the intermediate flow $F_{0 \rightarrow t}$ can be calculated according to Eq. \ref{eq1},\ref{eq2},\ref{eq3} and Fig. \ref{f1} as
\begin{equation}
F_{0 \rightarrow t}^x=\left\{
\begin{array}{ccl}
R(\cos{\theta_t}-\cos{\theta_0}) & & {|\sigma_{0 \rightarrow 1}| > \sigma_s}\\
t F_{0 \rightarrow 1}^x & & {|\sigma_{0 \rightarrow 1}| \leq \sigma_s}
\end{array} ,\right.
\end{equation}

\begin{equation}
F_{0 \rightarrow t}^y=\left\{
\begin{array}{ccl}
R(\sin{\theta_t}-\sin{\theta_0}) & & {|\sigma_{0 \rightarrow 1}| > \sigma_s}\\
t F_{0 \rightarrow 1}^y & & {|\sigma_{0 \rightarrow 1}| \leq \sigma_s}
\end{array} ,\right.
\end{equation}
where $\sigma_s$ means the threshold value and is set to $0.01$ in this paper. And $F_{1 \rightarrow t}$ can also be obtained similarly. Another thing to emphasize is that $R$ can be either positive or negative since $\sigma \in [-1,1]$. To make the sign meaningful, we let positive $R$ (or $\sigma$) indicate clockwise trajectory and negative indicate counter-clockwise, as shown in Fig. \ref{f1} (c).

\subsection{Forward Warping and Synthesis Network}
After getting the intermediate flows $F_{0 \rightarrow t}$ and $F_{1 \rightarrow t}$, forward warping is used to warp $I_0$ and $I_1$ to the temporal position $t$. We simply implement forward warping via average splatting $\overrightarrow{\Phi}$ introduced in SoftSplat \cite{soft2020}. Besides, we use a 3-level pyramid feature extractor $\psi(\cdot)$, which is the same as that in SoftSplat, to obtain the multi-scale features of both $I_0$ and $I_1$. Then these features are also warped, which is proven to bring significant improvements in the interpolation quality in \cite{soft2020}. Moreover, we additionally warp the curvature map $\sigma_{0 \rightarrow 1}$ and $\sigma_{1 \rightarrow 0}$ to implement curvature-based spatial attention in the synthesis network, which will be introduced next. The warping and synthesis process can be formulated as
\begin{equation}
\begin{split}
    &f_0^1,~f_0^2,~f_0^3=\psi(I_0),~~~~~~~~~f_1^1,~f_1^2,~f_1^3=\psi(I_1),\\
    &\widetilde{\sigma}_{0 \rightarrow 1},\widetilde{I}_0,\widetilde{f}_0^1,\widetilde{f}_0^2,\widetilde{f}_0^3=\overrightarrow{\Phi}([\sigma_{0 \rightarrow1},I_0,\psi(I_0)],F_{0 \rightarrow t}), \\
    &\widetilde{\sigma}_{1 \rightarrow 0},\widetilde{I}_1,\widetilde{f}_1^1,\widetilde{f}_1^2,\widetilde{f}_1^3=\overrightarrow{\Phi}([\sigma_{1 \rightarrow0},I_1,\psi(I_1)],F_{1 \rightarrow t}), \\
    &\hat{I_t}=g(\widetilde{\sigma}_{0 \rightarrow 1},\widetilde{\sigma}_{1 \rightarrow 0},\widetilde{I}_0,\widetilde{I}_1,\widetilde{f}_0^1,\widetilde{f}_1^1,\widetilde{f}_0^2,\widetilde{f}_1^2,\widetilde{f}_0^3,\widetilde{f}_1^3),
\end{split}
\end{equation}
where $~\widetilde{\cdot}~$ denotes the warped results and $g(\cdot)$ denotes the final synthesis network.

The synthesis network utilizes the warped results to generate the intermediate frame $\hat{I_t}$. Its detailed structure is depicted in the right part of Fig. \ref{f2}. Following \cite{soft2020,con2018}, we employ a modified version of GridNet \cite{grid2017} to avoid checkerboard artifacts. Differently, our model applies a GridNet with three rows and four columns. Besides, we add the curvature attention mechanism in the synthesis network, which is a type of spatial attention and takes in the concatenation of $\widetilde{\sigma}_{0 \rightarrow 1}$ and $\widetilde{\sigma}_{1 \rightarrow 0}$. The motivation is that pixels with high $\sigma$ values imply motion boundaries, hence the curvature attention can help to improve the quality of the synthesis frames.

\subsection{Loss Function}
Our training loss function is consisted of two terms. The first reconstruction term aims to reduce the absolute difference between the model output $\hat{I_t}$ and ground truth $I_t$
\begin{equation}
 \mathcal{L}_r = \|\hat{I_t}-I_t\|_1
\end{equation}
Following Liu \emph{et al.} \cite{dvf2017}, we use the robust generalized Charbonnier penalty function $\rho(x)=\sqrt{x^2+\epsilon^2}$ to optimize $\ell_1$ norm in the expression of $\mathcal{L}_r$, where $\epsilon=0.001$. 

The second term is the perceptual loss, which has been found to produce more visually realistic outputs \cite{per2016}. It is calculated with the feature extractor $\mathcal{F}$ from \texttt{conv4\_3} layer of the pretrained VGG16 network as follow
\begin{equation}
 \mathcal{L}_p = \|\mathcal{F}(\hat{I_t})-\mathcal{F}(I_t)\|_2
\end{equation}
Finally, the total loss function is $ \mathcal{L} = \mathcal{L}_r +\lambda \mathcal{L}_p$,
where $\lambda$ is the weight coefficient and we set $\lambda=0.01$ in the experiments.

\begin{figure*}[htb]
    \centering
    \includegraphics[width=17cm]{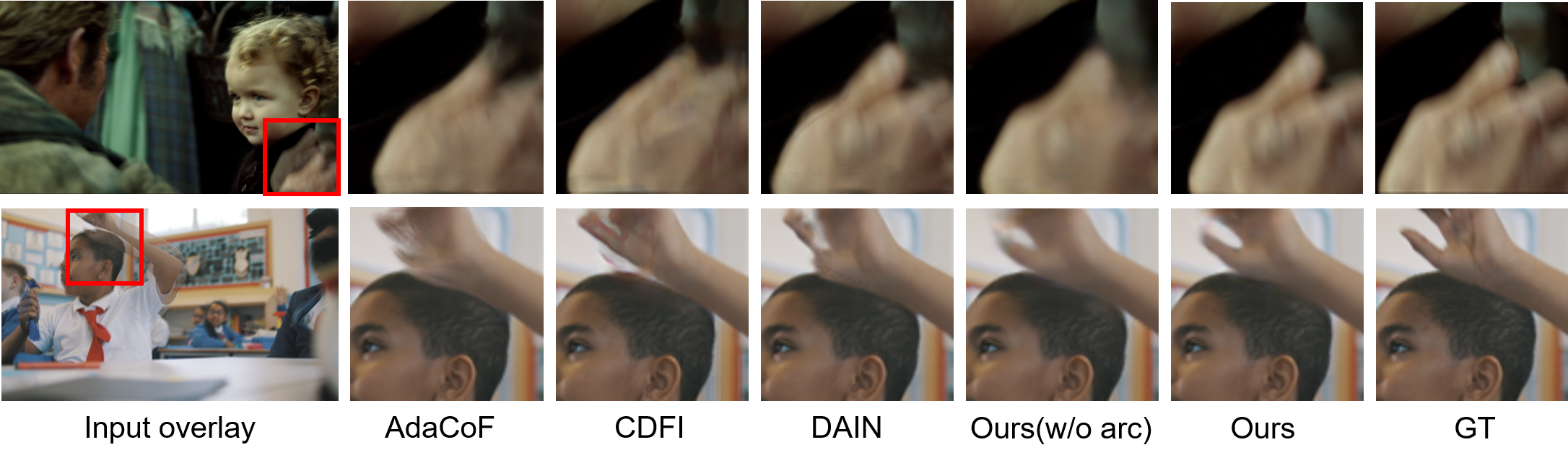}
    \caption{Qualitative comparison on Vimeo-90k \cite{xue2019video} testing set.}
    \label{f4}
\end{figure*}

\section{Experiments}
\subsection{Dataset, Metric and Training Strategy}
We train our model on Vimeo-90k dataset \cite{xue2019video}. It has 51,312 triplets for training and 3,782 triplets for testing, where each triplet contains three consecutive frames with a resolution of $448\times256$. We randomly augment the training data, including horizontal and vertical flipping, temporal order reversing, and rotating by 90 degrees. Note that the three frames in each triplet are equally spaced, which means we fix $t=0.5$. And we test our model on Vimeo-90k \cite{xue2019video}, UCF101 \cite{ucf101} and Middlebury \cite{mb2011} benchmarks. For quantitative evaluation, we measure the peak signal-to-noise ratio (PSNR), structural similarity (SSIM), and interpolation error (IE). Higher PSNR and SSIM scores and lower IE usually indicate better image quality. Our model is optimized by Adam \cite{adam2014} for 150 epoches on $256\times256$ patches from original frames and batch size is set to 8. The learning rate is initially set to 0.0002 and decays half every 30 epochs. All the experiments are conducted on two NVIDIA Tesla V100 GPUs.

\subsection{Ablation Studies}
To verify the effectiveness of our proposed ATCA model, we make several ablation studies. Specifically, We study on the effects of the two components: the arc trajectory and the curvature attention, hence we train another two networks without the two components respectively. Besides, we also train the complete ATCA without the perceptual loss, to verify the significance of this loss term. The study results are listed in Table \ref{t1}, which shows that the two components and the perceptual loss can lead to better performance. We also visualize the forward computation process of ATCA in Fig. \ref{f3}, including the bidirectional flow maps and $\sigma$-maps, intermediate flows and final output $\hat{I_t}$ with arc trajectories tagged on it. Note that brighter pixels in $\sigma$-maps have higher values, that correspond to the regions with large curvilinear motion in flow maps. This can explain why the curvature attention is helpful.

\subsection{Comparisons with SOTA Methods}
We compare ATCA with some previous SOTA approaches \cite{adacof2020,soft2020,dain2019,cdfi2021,bmbc2020}. Quantitative and Qualitative comparison results are shown in Table \ref{t2} and Fig. \ref{f4} respectively. Note that in SoftSplat method, we choose PWC-Net without fine-tuning as its flow estimator for fair comparison. We can find that our model performs better than these SOTA methods on Vimeo-90k and UCF101, and rank second on Middlebury. Moreover, ATCA has fewer parameters and relative fast speed. Obviously, our model have achieved high quality frames with clearer details and fewer artifacts as shown in Fig. \ref{f4}.

\begin{table}[tbp]   \small
\centering
\caption{\label{t1}Results of ablation studies.} 
\setlength{\tabcolsep}{0.7mm}{
\begin{tabular}{l|ccc}    
\toprule    
\multirow{2}{*}{Setting} & Vimeo-90k & UCF101 & Middlebury \\
\cmidrule(lr){4-4}\cmidrule(lr){2-2}\cmidrule(lr){3-3}
& PSNR & PSNR & IE \\
\midrule
ATCA(w/o arc trajectory) & 35.55 & 35.18 & 2.05 \\ 
ATCA(w/o curvature attention) & 35.75 & 35.20 & 2.04 \\
ATCA(w/o perceptual loss) & 35.71 & 35.19 & 2.02 \\   
\midrule
ATCA & \textbf{35.90} & \textbf{35.23} & \textbf{2.01}\\
\bottomrule   
\end{tabular}  }
\end{table}
\begin{table}[tbp]  \footnotesize
\centering
\caption{\label{t2}Quantitative comparison results with SOTA methods. To test the runtime, a pair of 640 × 480 images are processed by the models on one NVIDIA Tesla V100 GPU.}  
\setlength{\tabcolsep}{0.7mm}{
\begin{tabular}{lccccccc}    
\toprule    
\multirow{2}{*}{Method}&\multicolumn{2}{c}{Vimeo-90k}&\multicolumn{2}{c}{UCF101}& Middlebury & \#Params & Runtime \\
\cmidrule(lr){2-3}\cmidrule(lr){4-5}\cmidrule(lr){6-6}
& PSNR & SSIM & PSNR & SSIM & IE & (M) & (ms) \\
\midrule
SoftSplat \cite{soft2020} & 35.59 & 0.967 & - & - & - & 7.7 & - \\   
AdaCoF \cite{adacof2020} & 34.27 & 0.971 & 34.91 & 0.968 & 2.31 & 21.8 & \textbf{19}\\   
DAIN \cite{dain2019} & 34.71 & 0.976 & 35.00 & 0.968 & 2.04 & 24.0 & 356\\  
CDFI \cite{cdfi2021} & 35.17 & 0.977 & 35.21 & \textbf{0.969} & \textbf{1.98} & 5.0 & 124\\
BMBC \cite{bmbc2020} & 35.01 & 0.976 & 35.15 & \textbf{0.969} & 2.04 & 11.0 & 1257\\
\midrule
ATCA (ours) & \textbf{35.90} & \textbf{0.979} & \textbf{35.23} & \textbf{0.969} & 2.01 & \textbf{3.4} & 56\\
\bottomrule   
    \end{tabular}  }
\end{table}


\section{Conclusion}
In this paper, we propose a novel ATCA: an arc trajectory based model with curvature attention synthesis network, to synthesize video frames. Our model is trained on Vimeo-90k and tested on Vimeo-90k, UCF101 and Middlebury benchmarks. Ablation studies show that the proposed two components and the perceptual loss are effective in frame interpolation. Besides, our approach outperforms most of the SOTA methods and is also lightweight and fast.

\vfill\pagebreak
\bibliographystyle{IEEEbib}
\bibliography{strings,refs}

\end{document}